\documentclass[10pt,twocolumn,letterpaper]{article}

\usepackage{cvpr}
\usepackage{times}
\usepackage{epsfig}
\usepackage{graphicx}
\usepackage{amsmath}
\usepackage{amssymb}
\usepackage{adjustbox}

\usepackage[breaklinks=true,bookmarks=false]{hyperref}

\cvprfinalcopy %

\begin{document}

\title{Red blood cell image generation for data augmentation using Conditional Generative Adversarial Networks}

\author{Oleksandr Bailo, DongShik Ham, and Young Min Shin\\
Noul Inc.\\
{\tt\small alex,howard,young@noul.kr}
}

\maketitle

\begin{abstract}
In this paper, we describe how to apply image-to-image translation techniques to medical blood smear data to generate new data samples and meaningfully increase small datasets. Specifically, given the segmentation mask of the microscopy image, we are able to generate photorealistic images of blood cells which are further used alongside real data during the network training for segmentation and object detection tasks. This image data generation approach is based on conditional generative adversarial networks which have proven capabilities to high-quality image synthesis. In addition to synthesizing blood images, we synthesize segmentation mask as well which leads to a diverse variety of generated samples. The effectiveness of the technique is thoroughly analyzed and quantified through a number of experiments on a manually collected and annotated dataset of blood smear taken under a microscope.

\end{abstract}

\section{Introduction}

Deep learning based methods have had a great success on a number of typical visual perception tasks such as classification, segmentation, and object detection. While there are a number of important reasons that have made this progress possible, one of them is the availability of large-scale datasets. On the contrary to the traditional computer vision tasks, not much effort has been put to the creation of large-scale medical image datasets. There are several reasons these datasets are hard to create. 

First of all, there is a limited number of data annotators available. In contrast to traditional image annotation (e.g. class label, segmentation mask, bounding box) where fairly any person is able to perform the annotation, medical data requires a specialized professional\textemdash often with an advanced medical degree\textemdash to perform a reliable annotation. 

\begin{figure}
\begin{center}
\includegraphics[width=0.49\linewidth]{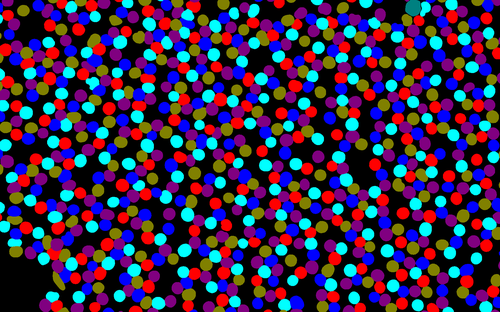}
\includegraphics[width=0.49\linewidth]{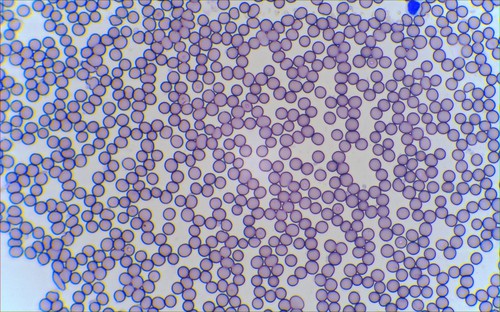}
\includegraphics[width=0.49\linewidth]{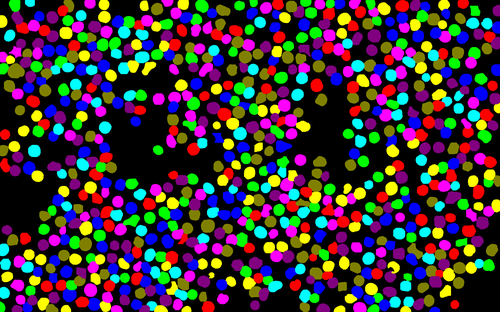}
\includegraphics[width=0.49\linewidth]{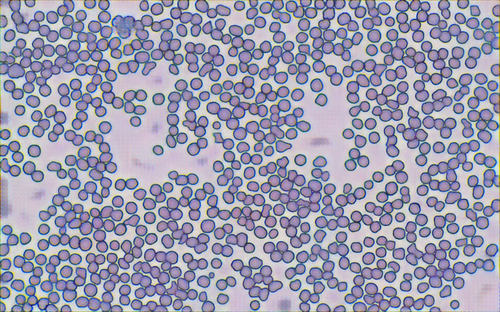}
\end{center}
    \caption{Image translation example. First row: real segmentation mask (left) and corresponding blood image (right). Second row: synthetic segmentation mask (left) and synthesized image (right).}
\label{Figure::TeaserImages}
\end{figure}

Secondly, medical data sharing is not a straightforward process. In order to democratize medical dataset, an agreement from a number of involved parties such as patients, doctors, hospitals, and data users should be reached. Additionally, it requires a detailed guideline on what grounds and for what purposes the data can be utilized.

Lastly, different hospitals and countries around the world might use different medical protocols, devices, or mechanisms. Thus, this data could not be easily transferable, requiring special managing to standardize it for the creation of a large-scale dataset.

\begin{figure*}
\begin{center}
\includegraphics[width=0.97\textwidth]{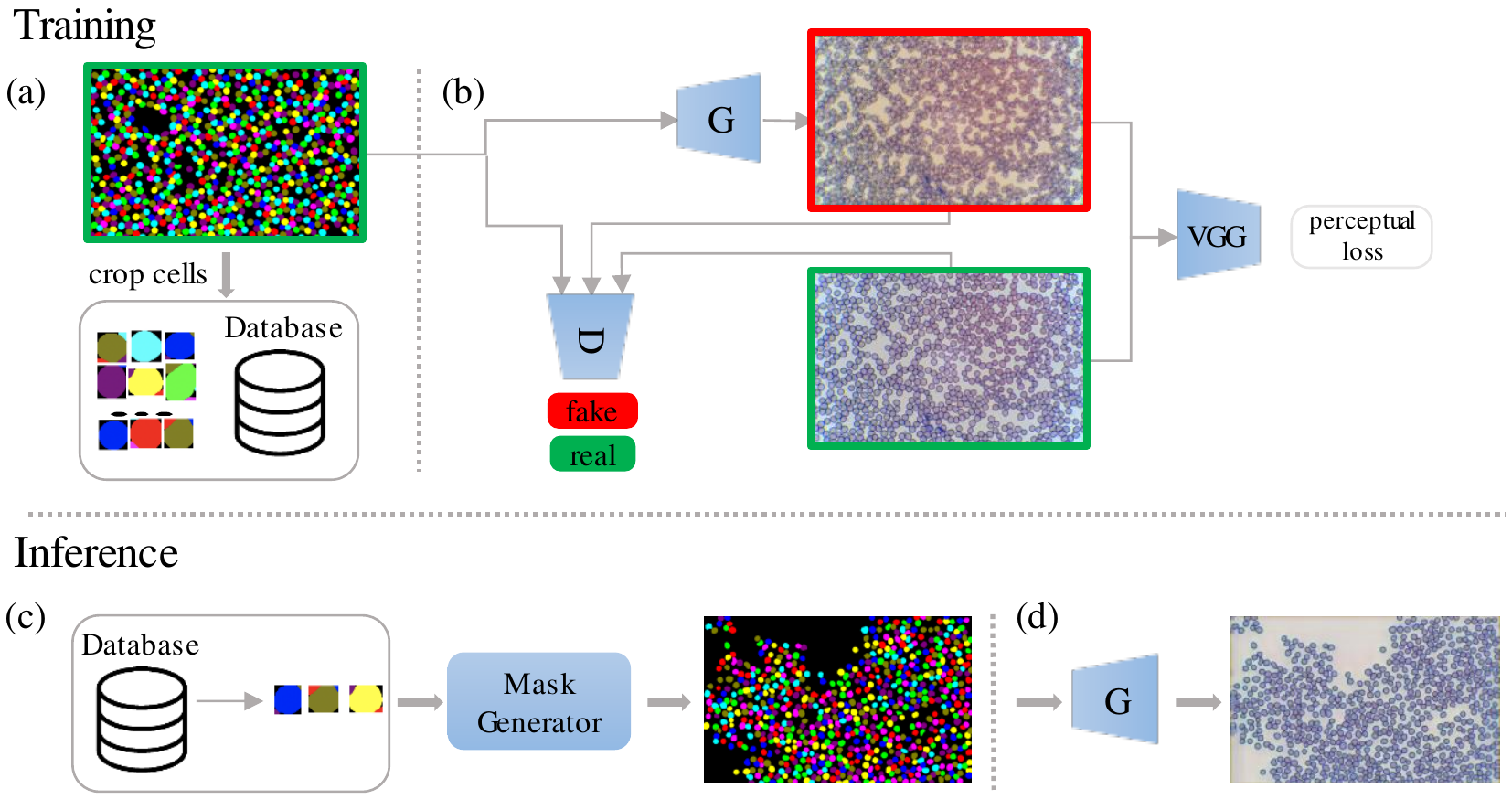}
\end{center}
    \caption{The overview of the proposed method. First, all blood cell shape instances are extracted and saved to a database (a). Meanwhile, pix2pixHD framework is trained to translate segmentation mask to blood cell images (b). During the inference stage, synthetic segmentation mask is created (c) and fed to the generator network to produce the realistic blood cell image (d).}
\label{Figure::MethodOverview}
\end{figure*}

All the of aforementioned reasons greatly increase the cost of the annotation while overall making a large-scale medical dataset creation extremely time and resource consuming. Fortunately, with the advent of the Generative Adversarial Networks (GANs)~\cite{goodfellow2014generative} a powerful image synthesis has become possible. Recently, Wang et al. have proposed a model named pix2pixHD~\cite{wang2018pix2pixHD} that performs high-resolution photorealistic image generation given the instance segmentation mask of the scene. Since the segmentation masks are easy to be manipulated, this method allows interactive object manipulation leading to numerous new samples.

In this paper, we focus on generating new training samples of blood images taken with a microscope. For this purpose, we have collected a dataset and manually annotated every single blood cell in images with an instance level segmentation label. Next, we train GAN to perform translation of a segmentation mask to a photorealistic blood image.

Once the network is trained, we automatically generate diverse instance segmentation masks on run (see Figure~\ref{Figure::TeaserImages}), and create a defined number of various data samples which are further used as an additional training data for segmentation and object detection tasks. These generated samples serve as a powerful data augmentation technique that could boost the performance of relevant tasks. The graphical summary of the algorithm is presented in Figure~\ref{Figure::MethodOverview}. To sum up, the contributions of this work are the following:
\begin{itemize}
    \item pipeline to generate red blood cell images from a synthetically created instance segmentation masks
    \item extensive quantified analysis on the effect of generated images on segmentation and detection tasks
\end{itemize}

This paper is structured in the following way. In Section~\ref{Section::RelatedWork}, we cover the related works which similarly have utilized GANs for medical data generation. Then, Section~\ref{Section::Dataset} describes the dataset creation. Section~\ref{Section::Methodology} specifies a procedure for a synthetic instance segmentation mask generation as well as details on the utilized GAN model. Lastly, a number of experiments and discussion are presented in Section~\ref{Section::Experiments} followed by conclusion and future plan (Section~\ref{Section::Conclusion}).

\section{Related Work}
\label{Section::RelatedWork}

The use of GANs with medical image data is not new~\cite{yi2018generative,kazeminia2018gans,wolterink2018generative}. GANs have demonstrated promising results with a medical data in a number of tasks such as segmentation~\cite{kohl2017adversarial,xue2018segan,moeskops2017adversarial,son2017retinal,yang2017automatic,zhang2017deep}, detection~\cite{schlegl2017unsupervised,baumgartner2017visual,chen2018unsupervised,sekuboyina2018btrfly,tuysuzoglu2018deep}, and image synthesis~\cite{galbusera2018exploring,wolterink2018blood,beers2018high,armanious2018medgan,nie2017medical}.

Since existing medical datasets are often limited in size, several works have examined the use of GANs for data augmentation purposes to increase the number of training samples. For instance, Calimeri et al.~\cite{calimeri2017biomedical} have concluded the possibility of employing synthetically generated MRI for inexpensive and fast data augmentation. Furthermore, Han et al.~\cite{han2019learning} have studied the effect of synthetic data on brain metastases detection task in MRI. In addition, several other works have observed an improved performance on segmentation of various data sources such as MRI~\cite{mok2018learning,shin2018medical}, CT~\cite{bowles2018gan}, and X-ray~\cite{neff2017generative}.

Similar conclusions of favorable use of GANs as data augmentation technique have been observed for classification of CT images~\cite{frid2018synthetic} and liver lesion classification~\cite{frid2018gan}.

\begin{figure*}
\begin{center}
\includegraphics[width=0.24\textwidth]{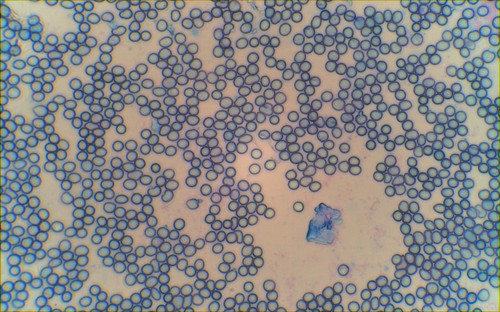}
\includegraphics[width=0.24\textwidth]{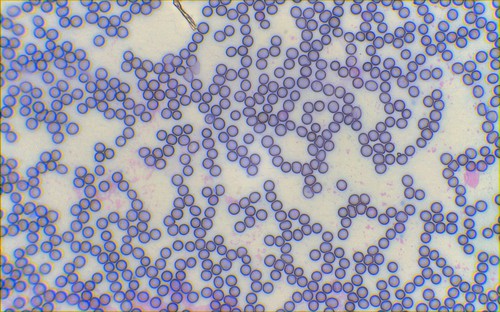}
\includegraphics[width=0.24\textwidth]{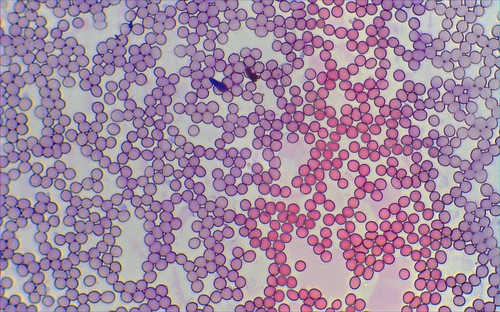}
\includegraphics[width=0.24\textwidth]{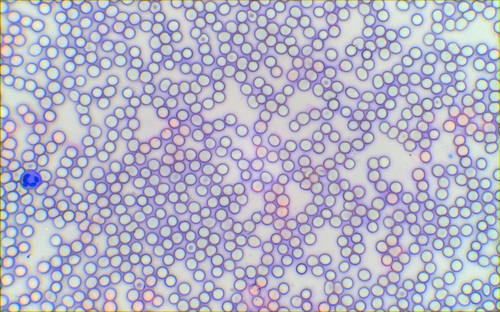}
\includegraphics[width=0.24\textwidth]{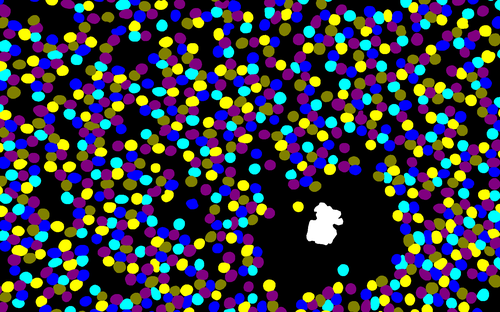}
\includegraphics[width=0.24\textwidth]{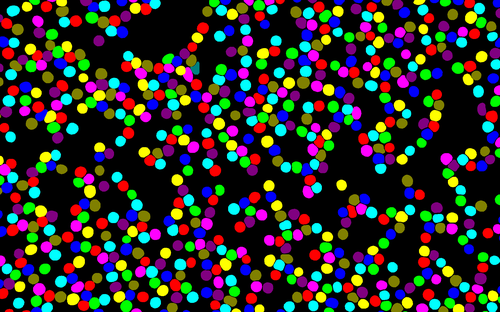}
\includegraphics[width=0.24\textwidth]{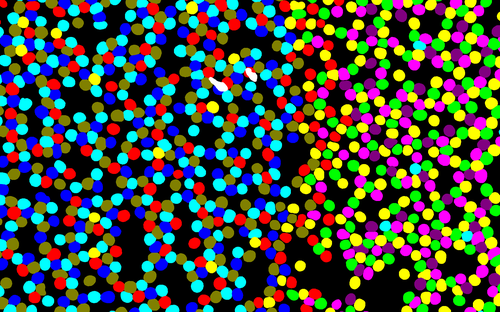}
\includegraphics[width=0.24\textwidth]{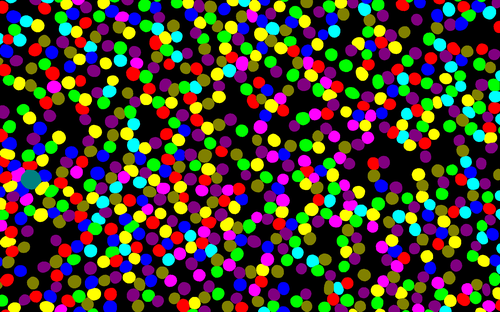}
\end{center}
    \caption{Samples from the dataset: blood cell images (first row) and corresponding instance segmentation masks (second row).}
\label{Figure::DatasetImamagesWithSegmentation}
\end{figure*}

While there are many works targeting MRI, CT, and X-ray image data, blood smear images taken with a microscope have got less attention. Thus, in this paper, we explore the use of GANs for blood image synthesis and study the effect of utilizing GANs as a data augmentation method for segmentation and object detection tasks.

\section{Dataset}
\label{Section::Dataset}

In order to obtain the dataset with Red Blood Cells (RBCs), we have collected blood from $100$ patients and prepared stained blood smear slides. Then, we have manually selected $100$ images to ensure every image is distinctive to diversify the dataset. This results in a great variation of a number of RBCs per image, their shape, and color values (see Table~\ref{Table::TrainSetStatistics}). For example, the largest cells could reach $71\times87$ resolution, while the smallest could be of several pixels (i.e. partially visible cells on the border of the image). Every image is of $1920\times1200$ resolution in RGB format. The dataset is randomly split into training and testing sets with a ratio of 60 to 40 respectively. Every image is of $1920\times1200$ resolution in RGB format.

The annotation technique of the dataset is inspired by the 2015 MICCAI Gland Segmentation Challenge~\cite{sirinukunwattana2017gland}. Specifically, given a microscope image with blood smear, we aim to create an instance segmentation mask which would allow us to extract every single cell in a given image. For this purpose, qualified professionals manually draw precise segmentation mask for each RBC. 

This drawing process consists in putting a color on every single pixel belonging to a blood cell. Color values are selected by the user from a predefined list of color values. In order to be able to extract every single blood cell without an overlap with others, any touching cells have different segmentation mask colors assigned to it, hence, no touching cells having the same color.

While this dataset also includes annotation for other blood cells and noise (e.g. dust), in this paper, we primarily focus on RBCs. The representative images and corresponding instance segmentation masks are shown in Figure~\ref{Figure::DatasetImamagesWithSegmentation}.

\begin{table}[]
\begin{tabular}{|l|c|c|c|c|}
\hline
                    & Mean & Std\\ \hline
RBCs per image & $669$     & $149$    \\ \hline
RBC shape           & $46\times46$    &  $5\times5$   \\ \hline
Colors (in RGB)     & $154\times151\times162$ &  $20\times17\times21$\\ \hline
\end{tabular}
\caption{Statistics of red blood cells in the train set.}
\label{Table::TrainSetStatistics}
\end{table}

\section{Methodology}
\label{Section::Methodology}

The creation of new samples is composed of two stages: synthetic mask generation and translating generated mask to a photorealistic image of blood cells. The graphical visualization of the whole pipeline during training and testing phases are shown in Figure~\ref{Figure::MethodOverview}.

\subsection{Synthetic mask generation}

In order to generate new and meaningful photorealistic samples of the blood cells, we first need to produce synthetic instance segmentation masks in which the blood cells have unique shapes and location distribution. For this purpose, we have designed a synthetic segmentation mask generator that combines sampled cell shapes and their distributions to produce synthetic segmentation masks.
More formally, we formulate a synthetic segmentation mask as a set of sampled cell shapes which are located at their corresponding locations on the background:
\begin{equation}
    \{ (s_1, l_1), (s_2, l_2), ..., (s_n, l_n), \text{background}\},
\end{equation}
where $s_i$ denotes the shape of the cell and $l_i$ determines the location of the cell. The total number of cells in an image $n$ is drawn from a normal distribution as $n \sim \textrm{Norm}(\mu_n, \sigma_n)$, where $\mu_n$ and $\sigma_n$ are determined from the statistics of training data (see Table~\ref{Table::TrainSetStatistics}).

\subsubsection{Cell shape sampling}

To model the natural variety and similarity of cell shapes in blood, we perform exemplar-based cell shape sampling.
During the training stage, segmented boundaries of each cell in the training data are accumulated to build the cell shape database. This database serves as a blood cell shape supplier during the inference stage. 
In the inference stage (i.e. generating new samples), the cell shape sampler iteratively selects random cell shape $s_i$ from the database and puts it on the instance segmentation mask which is composed of numerous cells.
Moreover, whenever we pick $s_i$ from the database, a set of probabilistic augmentations are applied to diversify the appearance. The augmentation includes rotation, scale, horizontal and vertical flipping.

\subsubsection{Cell distribution sampling}

The most straightforward way to locate the sampled cell shapes in the synthetic segmentation mask is simply to iteratively sample the coordinates at random and place the cell shape masks at the sampled location if the place is not occupied. However, such an approach results in a nonrealistic cell distribution. Specifically, this leads to a uniform random distribution of cells all over the image with many cells being placed solo without touching other cells. In reality, due to cell adhesion~\cite{gumbiner1996cell}, cells tend to stick to each other, consequently, forming clusters.

Therefore, in order to generate the segmentation masks which incorporate the aforementioned aspect, our cell distribution algorithm sequentially samples the appropriate location of each cell from the probability density function defined on the 2D discrete space.
The location of i-\textit{th} cell $l_i$ is sampled from the probability map $\mathcal{P}(i)$ as: 
\begin{equation}
    l_i \sim \mathcal{P}(i),
\end{equation}
where each pixel value in $\mathcal{P}(i)$ denotes the probability of being selected as a location of a cell's center at time step $i$.
$\mathcal{P}(i)$ is initially uniform during the sampling of the first $n_{init}$ cells, but changes its landscape as $i$ increases in order to simulate inherent cell adhesion. We have modeled this evolving nature of $\mathcal{P}(i)$ as a Markov random process.
\begin{align}
\label{Equation:markov}
    \mathcal{P}(i) = 
        \begin{cases}
            \mathrm{uniform}, & i \leq n_{init}\\
            \frac{1}{n_{init}}\sum_{j=1}^{n_{init}}z(l_j), & i = n_{init}+1\\
            (1-a_i)\mathcal{P}(i-1) + a_iz(l_{i-1}), & i > n_{init}+1
        \end{cases}
\end{align}
Specifically, this explains that the probability map $\mathcal{P}(i)$ incrementally changes as accumulating the function $z(\cdot)$ to its previous state. The excitation function $z(l_i)$ is calculated by applying a 2D Gaussian function with $\sigma=\sigma_x=\sigma_y$ around the sampled cell center $l_i$ and reverting the values within the $cell_{size}$ around it. This is done to impose low probability within boundaries of already allocated cells to prevent a cell being placed at the occupied location. The amount of increment is controlled by the normalizing coefficient $a_i$, which is in the form of a harmonic progression of $i$. (i.e. $a_i= 1/i$). At any time stamp, $\mathcal{P}(i)$ always maintains the `sum to unity' property.

In practice, during the cell placement on the synthetic mask, at any time the cell is placed on the canvas, a specific color is assigned to it to satisfy the condition that no touching cells have the same color. If such constraint is impossible to maintain\textemdash many touching cells or all predefined colors are used\textemdash the coordinate sampling is repeated. Since every touching cell has a different color, produced synthetic mask can be treated as an instance segmentation mask with a possibility to extract every single cell.

\begin{figure}
\begin{center}
\includegraphics[width=0.48\textwidth]{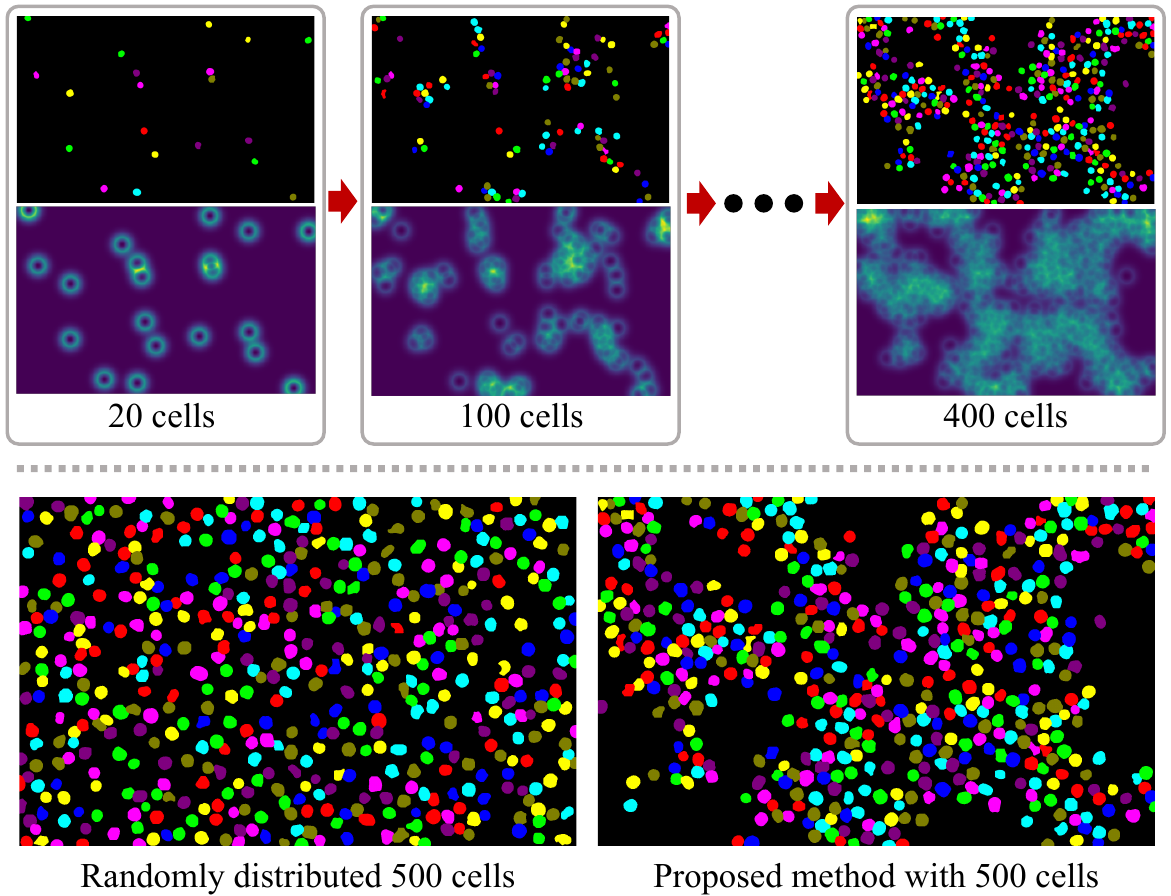}
\end{center}
    \caption{Synthetic mask generation. (Top) intermediate segmentation and probability maps of the algorithms. (Bottom) compares the final result of random placement and the proposed method.}
\label{Figure::SyntheticMaskGeneration}
\end{figure}

This synthetic mask generation process, as well as the comparison of randomly distributed cells against the proposed strategy, is visually described in Figure~\ref{Figure::SyntheticMaskGeneration}.

\subsection{Synthetic blood image generation}

For the purpose of synthesizing photorealistic blood cell image given the instance segmentation mask, we have utilized a recent pix2pixHD framework proposed by Wang et al.~\cite{wang2018pix2pixHD}. For our scenario, the pix2pixHD framework is composed of Generator $G$ which tries to translate segmentation mask to a photorealistic blood cell image. At the same time, two multi-scale generators $D = (D_1, D_2)$ are trying to distinguish real images from the generated ones. The full training objective of the network is the following:

\begin{equation}
\label{Equation::pix2pixHDTotalLossFunction}
\small
\begin{aligned}
    \min_{G}\bigg(\Big(\max_{D_1,D_2} \sum_{k=1,2}\mathcal{L}_{\text{GAN}}(G,D_k)\Big) + \lambda_{\text{FM}} \sum_{k=1,2}\mathcal{L}_{\text{FM}}(G,D_k)
    \\ + \lambda_{\text{PR}}~\mathcal{L}_{\text{PR}}(G(x, E(x)),y)\bigg),
\end{aligned}
\end{equation}

where:
\begin{itemize} 
\item $\mathcal{L}_{\text{GAN}}(G,D_k)$ is the adversarial loss defined as:

\begin{equation}
\mathbb{E}_{(x,y)}[\log D_{k}(x,y)]+ 
\mathbb{E}_{x}[\log (1- D_{k}(x,G(x, E(x)))],
\end{equation}

\item $\mathcal{L}_{\text{FM}}(G,D_k)$ is the feature matching loss that aims to stabilize training and produce more visually appealing results at multiple scales. It is defined as:
\begin{equation}
\small
    \mathbb{E}_{(x,y)} \sum_{i=1}^T \frac{1}{N_i}[||D_k^{(i)}(x,y)-D_k^{(i)}(x,G(x, E(x)))||_1],
\end{equation}

\item $\mathcal{L}_{\text{PR}}(G(x),y)$ is the perceptual reconstruction loss aiming to further improve the performance of high-quality image generation:
\begin{equation}
\small
    \sum_{i=1}^N\frac{1}{M_i}[||F^{(i)}(y)-F^{(i)}(G(x, E(x)))||_1]
\end{equation}

\end{itemize}

As suggested in~\cite{wang2018pix2pixHD}, we incorporate feature encoder network $E$ and combine its output with original input $x$ to be able to manipulate image synthesis style easily.  Since it is a relatively easy task to generate blood cells, we only use the global generator without a local enhancement.

During the inference stage, we feed synthetically generated mask to the generator to obtain a synthetic image of blood cells. The style of the output image is influenced by randomly sampling features from one of $10$ clusters which are created by running K-means clustering on the outputs of the encoder $E$ supplied with the training images.

For a detailed description of the losses, feature encoding, and clustering please refer to the original paper~\cite{wang2018pix2pixHD}. Implementation details, as well as training procedures, are described in Section~\ref{Section::pix2pixHDtraining}.

\section{Experiments and Results}
\label{Section::Experiments}

In this section, first, we cover specific details on parameters and training strategy for blood cell images synthesis. Later, the majority of this section describes various experiments on segmentation and object detection tasks with a focus on the effect of the use of synthetic images during training on the performance of the networks.

\subsection{Synthetic blood image generation}
\label{Section::pix2pixHDtraining}

In order to decide the number on cells to be placed on the synthetic mask, we sample the number from a normal distribution with a mean and standard deviation taken from the dataset statistics (see Table~\ref{Table::TrainSetStatistics} ``RBC count per image'' row). In the case of the probability map creation, during the synthetic mask generation, the standard deviation value $\sigma$ of 2D Gaussian distribution is related to half width at half maximum (HWHM)\footnote{It is half of the distance between points on the curve at which the function reaches half its maximum value.} value. Specifically, we want HWHM value to be equal to $cell_{size}$. Hence, we can derive that $\sigma=cell_{size}/\sqrt{2\ln{2}}$, where $cell_{size}=46$ (taken from Table~\ref{Table::TrainSetStatistics}). Initially, when a synthetic segmentation mask is empty, $n_{init}=20$ cells are placed at random without considering the probability map.

The training of the pix2pixHD is performed in two stages due to the GPU memory constraint. In the first stage, the images are downsized to $1024\times640$ and global generator $G$, discriminators $D$, and feature encoder $E$ are trained simultaneously for $120$ epochs. Then, feature encoder $E$ is used to precompute image features of the training data. In the second stage, the training of $G$ and $D$ is performed using full-resolution images (i.e. $1920\times1200$) for additional $120$ epochs. After training is completed, 10 clusters are created using encoded features for each semantic category. These clusters are used to simulate different blood cells style leading to the diversification of generated images.

In all experiments, we have utilized original pix2pixHD network with several modifications. For instance, the number of filters in the first convolutional layer of $G$, $D$, and $E$ are set to $16$. The primary goal for such channel tuning is to fit the model within the GPU memory capacity. Horizontal and vertical flipping are utilized as data augmentation.

\begin{figure}
\begin{center}
\includegraphics[width=0.49\linewidth]{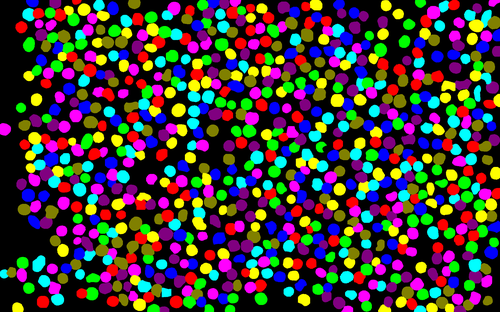}
\includegraphics[width=0.49\linewidth]{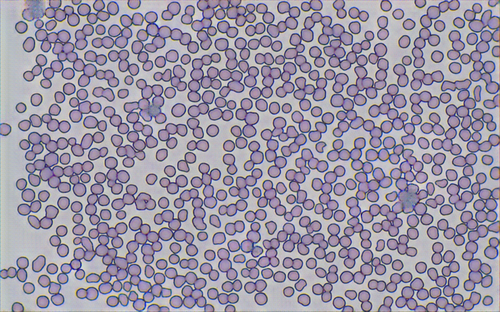}
\includegraphics[width=0.49\linewidth]{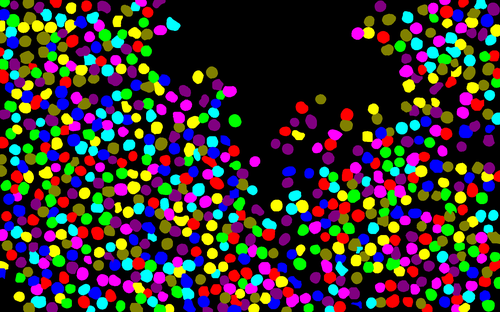}
\includegraphics[width=0.49\linewidth]{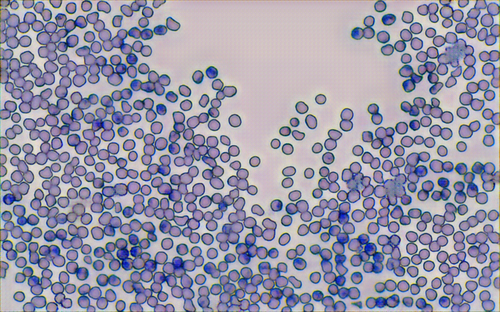}
\includegraphics[width=0.49\linewidth]{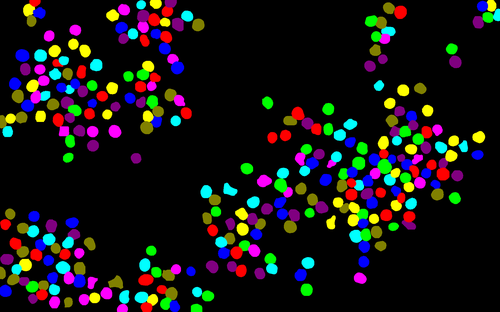}
\includegraphics[width=0.49\linewidth]{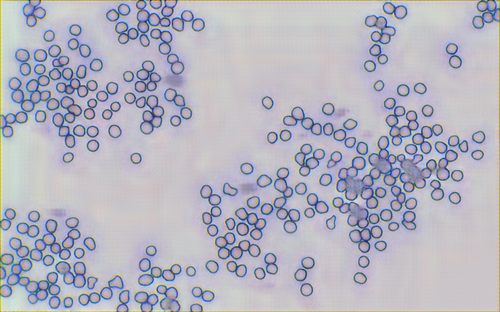}
\end{center}
    \caption{Examples of generated synthetic mask (left) and corresponding synthesized blood cell images (right).}
\label{Figure::GeneratedBloodImagesFromSynthetic}
\end{figure}

\begin{figure*}
\begin{center}
\includegraphics[width=0.24\textwidth]{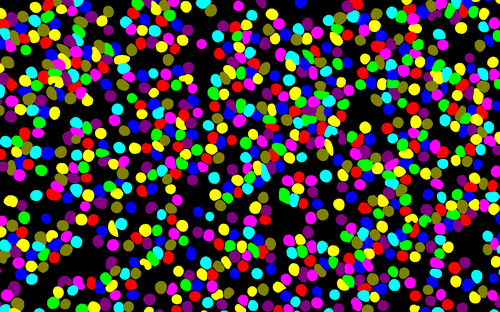}
\includegraphics[width=0.24\textwidth]{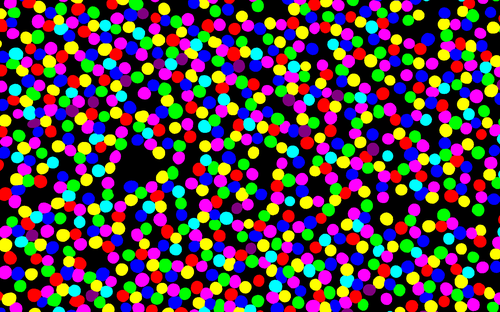}
\includegraphics[width=0.24\textwidth]{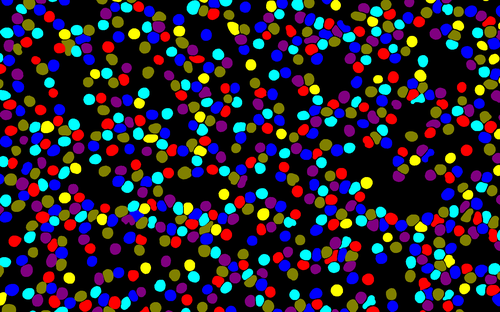}
\includegraphics[width=0.24\textwidth]{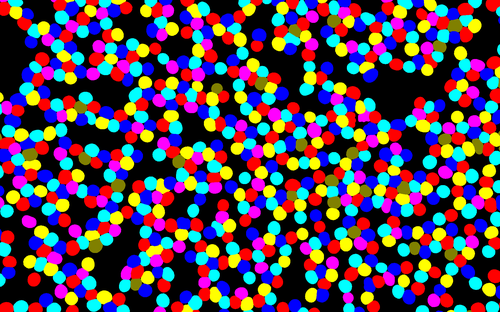}
\includegraphics[width=0.24\textwidth]{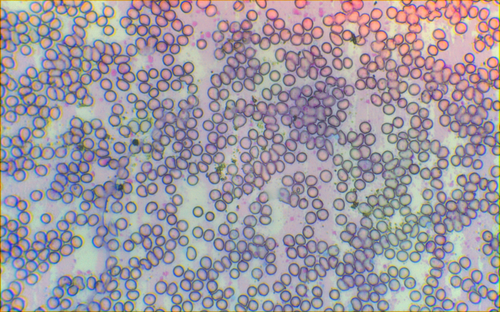}
\includegraphics[width=0.24\textwidth]{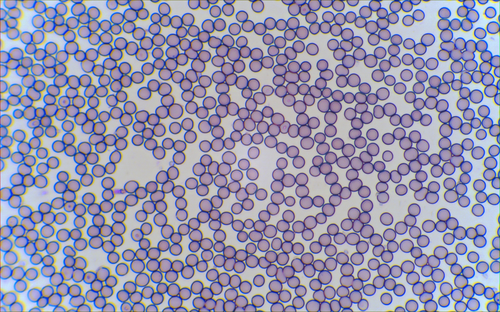}
\includegraphics[width=0.24\textwidth]{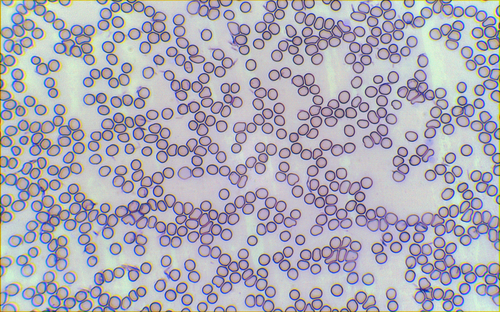}
\includegraphics[width=0.24\textwidth]{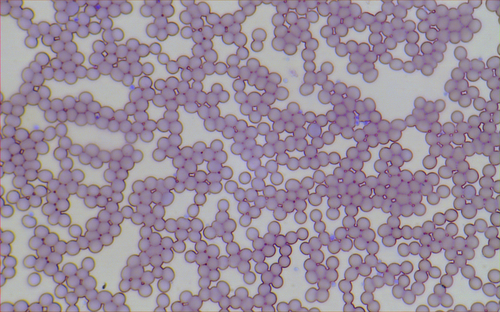}
\includegraphics[width=0.24\textwidth]{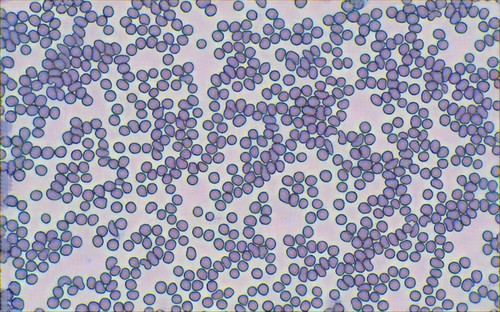}
\includegraphics[width=0.24\textwidth]{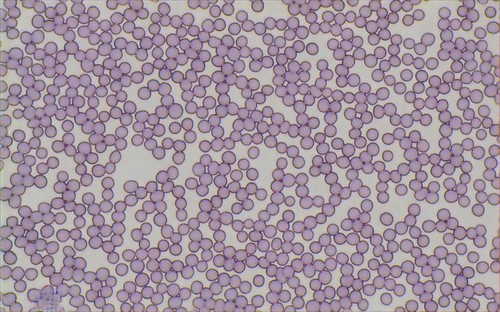}
\includegraphics[width=0.24\textwidth]{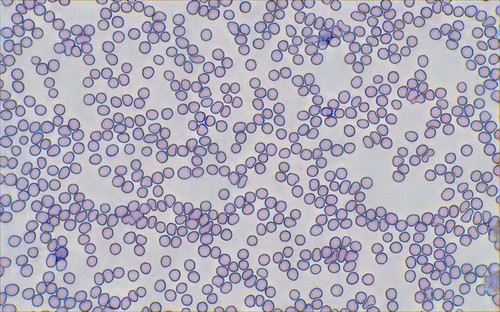}
\includegraphics[width=0.24\textwidth]{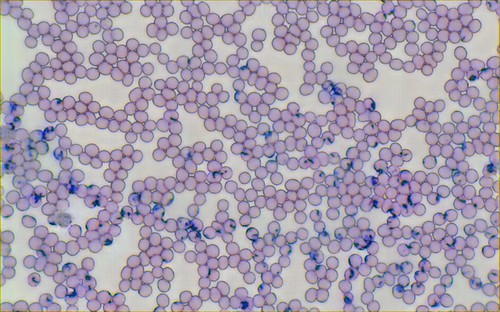}
\end{center}
    \caption{Generated samples on test set: segmentation masks (first row), ground truth (second row), and generated blood cells (third row).}
\label{Figure::GeneratedBloodImages}
\end{figure*}

\begin{figure*}
\begin{center}
\includegraphics[width=0.24\textwidth]{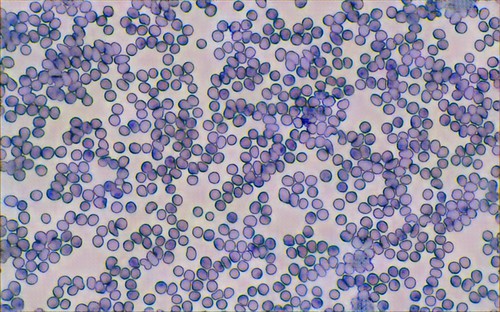}
\includegraphics[width=0.24\textwidth]{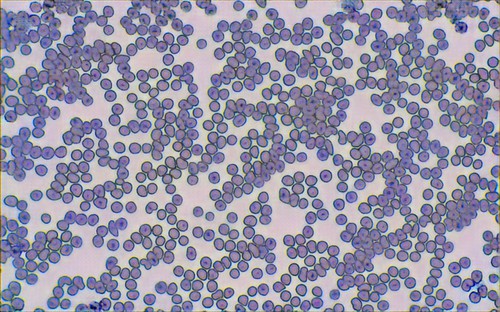}
\includegraphics[width=0.24\textwidth]{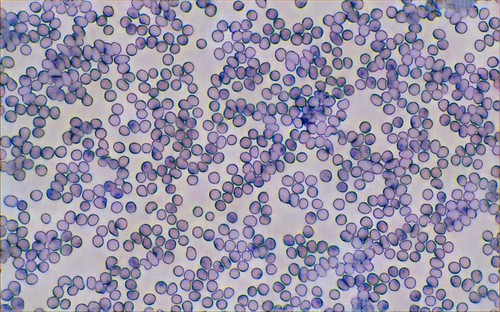}
\includegraphics[width=0.24\textwidth]{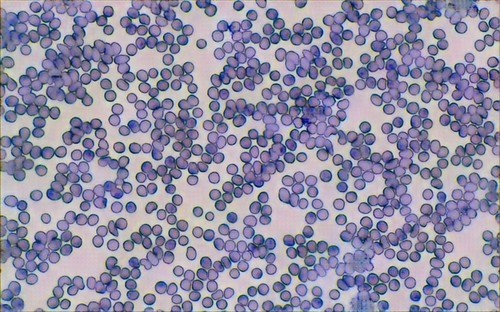}
\end{center}
    \caption{Example of different styles of the synthesized blood cell image.}
\label{Figure::GeneratedBloodImagesDifferentStyles}
\end{figure*}

The pix2pixHD network is trained with a train set solely. Figure~\ref{Figure::GeneratedBloodImages} shows network outputs on the test set images, hence, we can compare synthesized and corresponding real images. Certainly, the network is able to learn shapes, color, and boundaries of the blood cells given their segmentation mask and generally to produce realistic blood images. Noticeably, when synthetic and real images are compared one-to-one, they do not necessarily have the same colors and intensities values. Furthermore, due to the feature encoding which influences the style of individual cells, synthetic images often have excessive noize embedded to generated blood cells (see Figure~\ref{Figure::GeneratedBloodImages} (last column)). While the style is selected at random during the synthesis stage, we could manually define the cluster where features are sampled from. For example, Figure~\ref{Figure::GeneratedBloodImagesDifferentStyles} shows different styles of the generated blood cell image given the segmentation mask from the first column of the Figure~\ref{Figure::GeneratedBloodImages}.

The output results of the whole pipeline including synthetic mask generation and blood image synthesis can be seen in Figure~\ref{Figure::GeneratedBloodImagesFromSynthetic}. Noticeably, the proposed method is able to generate synthetic segmentation mask which holds cell adhesion rule (i.e. intercellular forces make cells group with each other). Furthermore, utilizing these synthetic segmentation masks as an input to the blood image generator, the results in realistic blood images of a different style (e.g. color and intensity values, noise level).

Unoptimized Python implementation of synthetic mask generation on average takes $60$ seconds (using Intel Xeon) per mask and it is heavily dependent on the number of cells to be placed on the mask. The image generation takes on average $0.5$ seconds per image (using GeForce 1080Ti). Therefore, in the current implementation form, proposed augmentation method better not be used for a real-time data augmentation, but rather for an offline technique to increase the training samples.

\subsection{Segmentation}
\label{Section::Experiments::Segmentation}

For this task, we have utilized FCN-8s~\cite{long2015fully} model. Loss function is formulated as $1 - DSC$, where $DSC$ stands for a Dice score that is defined as:
\begin{equation}
\small
    DSC= \frac{2|p\cdot t|}{|p|^2 + |t|^2},
\end{equation}
where it is stated as vector operation over binary vectors $p$ and $t$ corresponding to the network prediction and corresponding target (i.e. ground truth) respectively.

The Dice score is used as an evaluation metric as well. The model is trained from scratch without any pretraining or transfer learning until the training loss is converged and no improvement is observed on the test set. The best scoring weights on the test set are selected for reporting results.

\begin{table*}[]
\begin{center}
\begin{tabular}{|l|c|c|c|c|c|}
\hline
Methods \textbackslash Tasks                            & \multicolumn{5}{c|}{\textbf{Segmentation task (Section~\ref{Section::Experiments::Segmentation}). Metric: Dice score.}}    \\ \hline
                                                        & RD           & SD           & RD+aug        & RD+SD       & RD+SD+aug       \\ \hline
FCN~\cite{long2015fully}           & 0.961        & 0.848        & 0.962         & 0.948       & 0.964       \\ \hline
                                                        & \multicolumn{5}{c|}{\textbf{Detection task (Section~\ref{Section::Experiments::Detection}). Metric: AP.}}              \\ \hline
Faster R-CNN~\cite{ren2015faster} & 0.781        & 0.852        & 0.985         & 0.986       & 0.993           \\ \hline
                                                        & \multicolumn{5}{c|}{\textbf{Detection from segmentation (Section~\ref{Section::Experiments::DetectionFromSegmentation}). Metric: AP.}} \\ \hline
DCAN~\cite{chen2016dcan}          & 0.853        & 0.749        & 0.885         & 0.848       & 0.895           \\ \hline
\end{tabular}
\end{center}
\caption{Quantitative evaluation on various tasks. Column names represent a different combination of data used for training of the networks: real data(RD), synthetic data(SD), and augmentations (aug).}
\label{Table::QuantitativeResults}
\end{table*}

Table~\ref{Table::QuantitativeResults} demonstrates the quantitative results with different data used as training instances. For instance, ``RD'' column represents results where the network is only trained with real data (RD). Similarly, in the ``SD'' case, exclusively synthetic data (SD) is used for the training. In both of these cases, the number of training samples is intentionally set equal to $60$ samples. The remaining columns ``RD+aug'', ``RD+SD'', and ``RD+SD+aug'' represent results where training set is composed of real data with augmentations, real data mixed with synthetic data, and augmented real and synthetic data respectively. The data augmentation includes horizontal and vertical flips, Gaussian blur, sharpening, embossing, Gaussian noise, color channel inversion, brightness change (addition and multiplication), contrast normalization, and grayscale augmentation applied randomly. For these scenarios, the number of training samples is identical and equals $960$ samples. 

FCN is a powerful model capable of reaching a high dice score of $0.961$  by utilizing real data exclusively without any augmentations. Though, when the real data is replaced with synthetic one, the network reaches a substantially lower score of $0.848$. This could imply that the real data has richer semantic information compared to the completely synthetic images that tend to approximate the real data.

Real data augmentation slightly helps to boost performance by about $0.1$ percent points. However, utilizing synthetic data as an augmentation method alongside real data actually harms the performance of the model decreasing the score by roughly $1\%$. Lastly, if the real and synthetic data is trained with augmentations together, the maximum score of $0.964$ is reached.

To sum up, this experiment implies that the use of generated synthetic blood cell images is beneficial to the performance of the network in this particular circumstances. The similar conclusions have been observed in the previous works~\cite{mok2018learning,shin2018medical,bowles2018gan,neff2017generative} as well.

\subsection{Detection}
\label{Section::Experiments::Detection}

For this task we have utilized Faster R-CNN~\cite{ren2015faster} based on ResNet-101~\cite{he2016deep} feature extractor. The model is pretrained on
Visual Genome~\cite{krishnavisualgenome}, and we have fine-tuned the model with blood cell data sets mentioned in Table~\ref{Table::QuantitativeResults}. For this task average precision (AP) is used as an evaluation matric. The implementation of the network is provided by~\cite{jjfaster2rcnn}.

Faster R-CNN is a heavy network which requires lots of training data. Hence, utilizing only limited real data for training results in a relatively low AP score of 0.781 (see Table~\ref{Table::QuantitativeResults}). Surprisingly, the network performs better (about $9\%$ relative improvement) when synthetic data alone is used for training while every other factor is kept equal. While this is counter-intuitive, it could be explained by the fact that, in order to produce synthetic blood images, we produce synthetic segmentation masks which comprise of blood cells shapes derived from a whole training set. This results in new cell shape distribution for each synthetic image. Furthermore, when the cell shape is placed on the segmentation mask, we additionally apply augmentations for each cell shape which leads to even more diverse samples compared to the real training set.

Contrarily to the segmentation task, augmentation provides a drastic improvement (about $26\%$) to the generalization capabilities of the model for the cell detection task. Similarly, a noticeable improvement is observed if synthetic data is used alongside the real data. Lastly, using real and synthetic data with augmentations provides even further improvement (about $1\%$) to the performance compared to the one without synthetic images.

Overall, this experiment shows a marginal benefit of synthetic data on the detection task when used alongside real data. Utilization of synthetic data during the training performs on par with traditional data augmentation techniques and could be used alongside real data and augmentations to maximize the model performance.

\subsection{Detection from segmentation}
\label{Section::Experiments::DetectionFromSegmentation}

While Faster R-CNN model has shown a great performance on the cell detection task, this model has several significant obstacles. First of all, it struggles with cluttered objects which highly overlap with each other. This is a potentially difficult problem because it involves a large number of small highly-overlapping objects (i.e. small object detection problem). Secondly, such a complex object detection models are often quite computationally expensive, which might be a problem for point-of-care medical devices which often require a fast and efficient computation on the edge. Lastly, this model is relatively slow and hard to stably train with numerous parameter tuning required.

In order to implement a method which is light-weight and fast, while maintaining satisfactory detection score, we utilize an idea proposed by Chen et al.~\cite{chen2016dcan}. Specifically, the segmentation network is modified to predict objectness (i.e traditional segmentation mask) as well as contours (i.e. boundary) of the blood cells. These two predictions are further processed to isolate individual cells from each other which simplify the detection problem to a blob detection on the binary image. Therefore, this method is suitable for the detection task and can be evaluated with average precision (AP) score as in the previous experiment.

Similarly to the segmentation task described in Section~\ref{Section::Experiments::Segmentation}, we have utilized FCN network but with the modified head to output objectness and contour predictions. The network is trained from scratch, with a loss which is simply a sum of objectness and contour losses. The best performing (defined by the AP score) model on a test set is selected for results stated in this paper.

The evaluation results are stated in Table~\ref{Table::QuantitativeResults}. Since this method relies on segmentation, the detection performance generally follows the trend of segmentation prediction from Section~\ref{Section::Experiments::Segmentation}. For example, utilizing only synthetic data harms the performance of the network by $0.104$ AP score. Also, the use of data augmentation helps to marginally boost to the network performance. Finally, using real and synthetic data with augmentations helps to reach the highest AP score of $0.895$. When this approach is compared to Faster R-CNN model, only in the case the real data is used for training, this method is able to achieve better performance. However, in every other scenario, Faster R-CNN outperforms this approach by a large margin.

While with this approach we have not succeeded to outperform powerful Faster R-CNN model, we confirm the possibility of achieving blood cell detection from the segmentation mask. The quality of detection is directly related to the quality of predicted segmentation mask. Therefore, we believe that the detection performance could be greatly increased by utilizing more recent neural networks such as~\cite{kayalibay2017cnn} for more accurate medical image segmentation.

\section{Conclusion}
\label{Section::Conclusion}

In this paper, we have developed a method to synthesize photorealistic microscopy blood images by utilizing conditional generative adversarial networks. These synthetic images are used alongside real data to meaningfully increase small datasets. The effect of such data augmentation technique is studied through a number of experiments on several tasks. While the use of synthetic images is shown to be marginally beneficial for the segmentation task, the performance on a detection task demonstrates a slightly stronger relative improvement.

To sum up, based on the experiments we have performed, for our specific strategy and algorithm design, the use of GANs as a synthetic data generator and further utilization of generated samples as an augmentation technique is usually beneficial for the model performance. However, the additional overhead which comes from designing GAN model, long and unstable training, heavy computational requirements, and other challenges might not justify the marginal improvement to the overall performance for a specific task.

In the current version, the proposed method is limited to generating microscopy red blood cell images. In future work, we plan to extend this method to synthesize other blood cells such as white blood cells, and platelets. Additionally, we would like to be able to synthesize parasite-infected cells \textemdash red blood cells with a visible parasite inside them \textemdash which would be beneficial for identification of various diseases such as malaria.

{\small
\bibliographystyle{ieee}
\bibliography{egpaper_final.bbl}
}

\end{document}